\documentclass{acm_proc_article-sp}

\newcommand{\UPLB}{University of the Philippines Los Ba\~{n}os}

\let\footnotesize\small

\begin{document}

\title{Capturing the Dynamics of Pedestrian Traffic\\Using a Machine Vision System}
\numberofauthors{1}
\author{
\alignauthor Louie Vincent A. Ngoho and Jaderick P. Pabico*\\
   \affaddr{Institute of Computer Science}\\
   \affaddr{\UPLB}\\
   \affaddr{College 4031, Laguna}\\
   \email{*jppabico@uplb.edu.ph}
}
\date{}
\maketitle
\begin{abstract}
We developed a machine vision system to automatically capture the dynamics of pedestrians under four different traffic scenarios. By considering the overhead view of each pedestrian as a digital object, the system processes the image sequences to track the pedestrians. Considering the perspective effect of the camera lens and the projected area of the hallway at the top-view scene, the distance of each tracked object from its original position to its current position is approximated every video frame. Using the approximated distance and the video frame rate (30 frames per second), the respective velocity and acceleration of each tracked object are later derived. The quantified motion characteristics of the pedestrians are displayed by the system through 2-dimensional graphs of the kinematics of motion. The system also outputs video images of the pedestrians with superimposed markers for tracking. These visual markers were used to visually describe and quantify the behavior of the pedestrians under different traffic scenarios.
	
\end{abstract}


\section{Introduction}
Pedestrian traffic is the behavioral flow of people in an area or moving crowd. In any publicly-accessible area, variations of pedestrian traffic can always be observed. There are several places in these public areas where the sudden change in the behavior of huge groups of people can cause traffic jams, specifically along hallways and on crossroads. The measurement of the dynamics of pedestrian traffic is important in various applications and field of studies such as crowd control, stampede and accident control, public security, pedestrian flow organization, pedestrian safety, architectural design and safety, advertising, and marketing. For example, architects and transportation designers frequently use crowd dynamics in designing environments and buildings for the efficiency of pedestrian flow. During extreme conditions, such as stampedes, riots, and fire escapes, improved procedures on building designs could save many lives~\cite{ref1}.
	
In obtaining data about the dynamics of pedestrian traffic, the usual procedure is to manually observe the various metrics associated with particle motion of a pedestrian by assigning several researchers on top of walkways over a period of time. This manual procedure is tedious, prone to error, and sometimes unreliable because of the human factors affecting the observers. During the start of the observation, the data gathered might be accurate because the human observers are still attentive. However, towards the later part of the observation, the data may become inaccurate due to the observers' boredom, eye stress, and tiredness. Thus, to replace the human observers, we created a computer vision system to mimic the observing capabilities of humans at a higher accuracy and consistency during the longest possible time. This study aims to effectively present the pedestrian behavior and characterize the motion of the crowd using the developed vision-based system.
	
Several researchers have attempted to simulate the pedestrian traffic dynamics wherein models for crowd dynamics were developed~\cite{ref1}. Some of these models are the agent-based model of Reynolds~\cite{ref4}, the social force model of Helbing and Molnar~\cite{ref6}, the cellular automata model of Blue~\cite{ref3}, and the gas-kinetic model of Helbing~\cite{ref5}, but all of them used data gathered by human observers. To develop better models, data must be gathered by a system that is less prone to error due to human subjectivity and tiredness, and thus, a vision-based system needs to be developed. To develop our vision-based system, we utilized models, concepts, and techniques in image processing and computer vision. Our aim for the system is to detect and track pedestrians and at the same time measure and analyze the pedestrian's collective behavior under four different traffic scenarios. The system accepts as input the overhead view of the pedestrian traffic. To aid the system operator in real-time, color is used to track multiple pedestrians. The system, then, outputs several metrics and captures the dynamics of these metrics in 2-dimensional graphs of the kinematics of motion: distance vs. time, velocity vs. time, and acceleration vs. time.

\section{Related Literature}
\subsection{Recent Studies in Traffic Dynamics}
Because of the importance of crowd dynamics on several real-world applications, several researchers attempted to quantify the collective dynamics of the pedestrians through developed simulations of motion. Helbing and Molnar~\cite{ref6}, borrowing some ideas in gas-kinematic models, introduced the social force model to simulate pedestrian flows.  In their model, a self-driven particle (i.e., a pedestrian) that interacts through social rules and regulations tries to move in its desired speed and direction while at the same time attempts not to collide with obstacles, other particles, and surrounding barriers.  In order to reach its destination faster, pedestrians take detours even if the route is crowded~\cite{ref8}. The choice, however, is dependent on the recent memory of what the traffic was like the last time they took the route, which was found by other researcher to be polygonal in nature~\cite{ref9}. In agreement with the social force model, Weidmann~\cite{ref10} observed that, as long as it is not necessary to go faster, such as going down a ramp, a pedestrian prefers to walk with his or her desired speed, which corresponds to the most comfortable walking speed. However, Weidmann further observed that pedestrians keep a certain distance from other pedestrians and borders. The distance between the pedestrians decreases as the density of the crowd increases. The pedestrians themselves cause delays and obstructions. Arns~\cite{ref12} observed that the motion of the crowd is similar to the motion of gases and fluid, while Helbing, {\em et al.}~\cite{ref8} suggested that it is similar to granular flow as well.

Helbing~\cite{ref11}, in his extension of the social force model, showed that many aspects of traffic flow can be reflected by self-driven many-particle systems. In this system, he identified the various factors that govern the dynamics of the particles such as the specification of the desired velocities and directions of motion, the geometry of the boundary profiles, the heterogeneity among the particles, and the degree of fluctuations. One such observable pattern is the formation of lanes of uniform walking direction, formed because of the self-organization of the pedestrians~\cite{ref13}. Aside from the self-organizing behavior of the crowd, obstacles were also observed to both positively and negatively contribute to the flow of the traffic. 

During escape panic of large crowds, several behavioral phenomena were observed~\cite{ref2}: build up of pressure, clogging effects at bottlenecks, jamming at room widening areas, {\em faster-is-slower effect}, inefficient use of alternative exits, initiation of panics by counter flows, and impatience. It was observed that the main contributing behaviors in these situations is a mixture of individual and grouping behavior.

\subsection{Tracking and Detecting Objects}
Detection and tracking of objects using machine vision is a much efficient solution, with much higher consistency of results, than the canonical manual observation performed by multiple human observers. Several methods have been implemented and tested in the area of machine-based tracking, the most traditional of which are the background subtraction method and the frame differencing method. However, the most efficient approach~\cite{ref15} by far is by utilizing a bipartite graph~$G^\beta$. In this method, $G^\beta$ contains  classes that represent certain entities about a frame or image parsed from a video. One class corresponds to the expected positions of the tracked objects~$E^p_i\in B, \forall i$, while the other corresponds to the blobs~$B$ from within the current image. Features such as size, position, color, shape, and velocity may be extracted from the object blobs. An adjacency matrix~$M_G$ of the graph will be created, wherein the matrix entries $m_{i,j}$ are the similarity measurements between the $j$th blob and $i$th tracked object.

\subsection{Motion Analysis}

Researches abound in the study of motion in image sequences. {\em Motion analysis is the science of comparing sequential still images captured from photographing a body in motion in order to study the kinematics (the motion themselves) and the kinetics (external and internal forces)} of the body~\cite{ref17}. Often called dynamic image analysis~\cite{ref16}, motion analysis is only based on a few number of consecutive frames, which has similar methods as to the analysis of static images. Correspondence between the pairs of points of interest are being observed in the sequence of images. It uses the concept of motion field which projects the three-dimensional motion into two-dimensional one, such that a velocity vector (e.g., direction, velocity, and distance) from an observer is assigned to each point in the two-dimensional image. A good approximation of the motion field is the concept of the optical flow. Optical flow gives a description of motion and can be a valuable contribution to the interpretation of the image. Since the data is qualitative, optical flow can be done even if no quantitative data are obtained from motion analysis. Furthermore, optical flow can be used to study a large variety of motions, as it can determine the motion direction and velocity at image points. The method, however, suffers in accuracy  due to illumination changes but can still be useful by taking in some motion-localized assumptions, such as constraining the motion to maximum velocity, small acceleration, common motion, and mutual correspondence.

\section{Method}

\begin{figure}
		\centering
			\epsfig{file=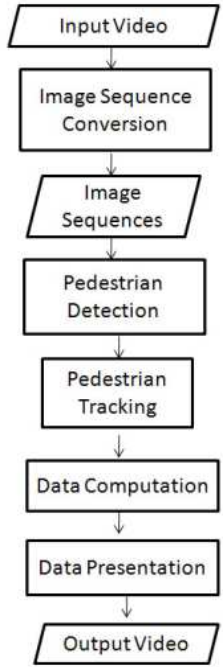,width=1.25in}
		\caption{Flowchart of the system showing major processes.}
		\label{Figure 1}
\end{figure}	

\subsection{Video Capturing}
We captured several videos in overhead view of pedestrians along hallways and cross ways under different traffic scenarios as follows:
\begin{enumerate}
  \item {\bf Scenario 1}: Pedestrians were flowing smoothly in one direction (left-to-right, LR).
	\item {\bf Scenario 2}: Pedestrians were flowing in two opposite directions simultaneously (LR and right-to-left, RL).
	\item {\bf Scenario 3}: Pedestrians were initially flowing from a single direction (LR) and then suddenly, a pedestrian stops in the middle and makes a U-turn.
	\item {\bf Scenario 4}: Pedestrians were moving against counter flows: They were passing by each other in several and inconsistent pathways or lanes.
\end{enumerate}

\begin{figure}[htb]
		\centering
			\epsfig{file=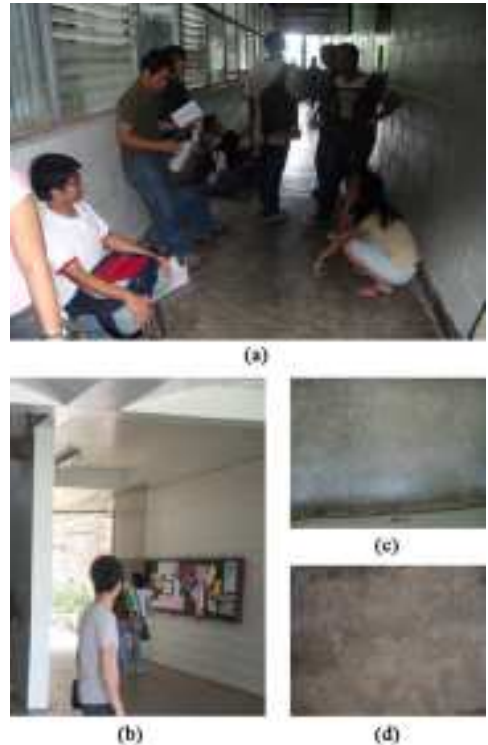,width=2.5in}
		\caption{Study area (a) corridor, (b) lobby, (c) overhead view of the corridor, and (d) overhead view of the lobby. This figure is in color in the electronic copy of this paper.}
		\label{Figure 2}
	\end{figure}

In the corridor of the study area (Figure~\ref{Figure 2}~(a)), after considering for the perspective effect of the camera lens, the dimensions of the captured area (Figure~\ref{Figure 2}~(c)) is $2.0\times 1.5$ (3 m$^2$ ) when the camera was mounted 3.5 m from the hallway floor (Figure~\ref{Figure 2}~(b)). However, the captured area (Figure~\ref{Figure 2}~(d)) is 4.5 m$^2$ ($2.5\times 1.8$) when the camera is mounted 3.9 m from the ground. We used a video rate of 30 frames per second (fps).
	
\subsection{Preprocessing Image Sequences} 
We split into several frames $F = \{f_1, f_2, \dots\, f_n\}$ the respective $n$-frame videos of the different pedestrian scenarios using a video processing software called VirtualDub (Version 1.6.16)~\cite{virtualdub}. Because of the variability of the scenarios, each has a different value for~$n$. We then archived the image frames in file directories.

\subsection{Detecting Pedestrians}
\begin{figure}[htb]
		\centering
			\epsfig{file=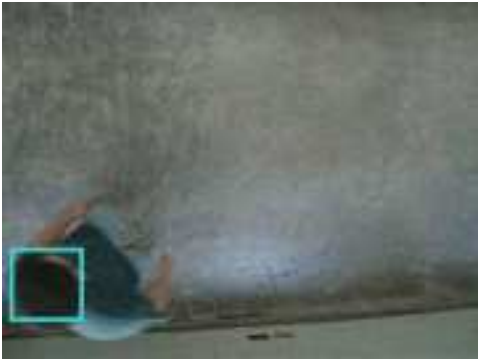,width=2.5in}
		\caption{An image frame showing an example of a pedestrian's head as detected by the procedure and bounded by~$b$. This figure is in color in the electronic copy of this paper.}
		\label{fig:detect}
	\end{figure}

We used the hair color of the pedestrians to detect them by utilizing various intensities of the black pixels. We did this by scanning each $f_i, \forall i=1,2,\dots,n$ through a bounding box~$b$, and then by processing the two-dimensional pixels bounded by~$b$. Within each~$b$, we counted the valid red-green-blue (RGB) color values of black pixels. We indicated that~$b$ has already bounded a candidate pedestrian head once a threshold sum of pixels~$\sum p$ is met (See Figure~\ref{fig:detect} for an example overhead image with detected head). We then searched the vicinity margin of~$b$ for more related pixels. Once done, we incremented the position of~$b$ and repeated the same process for the next set of pixels bounded by~$b$.
	
\subsection{Tracking Pedestrians}
Once we have identified that~$b$ bounds a pedestrian's head, we computed the two-dimensional coordinates of the center of mass $M_{x,y}(O)$ of the bounded object~$O$. We used $M_{x,y}$ to track the coordinates of the center of mass of the pedestrian's head. We considered~$b$ to be still bounding the pedestrian's  head if the difference of the current $(x_{t+1},y_{t+1})$ and previous $(x_t,y_t)$ of $M(O)$ is less than a set value (See Table~\ref{Table I} for the set of values we used in this study). Otherwise, we considered the~$f$ as erroneous and used $M_{x,y}(O)$ at time~$t$ instead. Because the pedestrians head might overlap in real-world due to the perspective effect of the camera lens, we allowed~$b$ to overlap with other~$b$s in a certain margin (See column F of Table~\ref{Table I} for the margin values we used in this study). We repeated the process for the succeeding~$f$ until the object being tracked is out of the scene. We considered an object as out of the scene when there is no more valid number of black pixels within~$b$ that can represent the object.

We utilized several threshold values to account for the varied lighting conditions when the input image sequences were recorded for each different scenario. We did this to avoid false detection and misdetection of pedestrians in the image sequences. The values of variables we used for the different scenarios are shown in Table~\ref{Table I}.
	
\begin{table}[htb]
\caption{Different values of variables for different scenarios. A - the dimensions of~$b$; B - Maximum RGB color value of black pixel for the object; C - Minimum number of valid black pixels; D - Vicinity margin of the object  (in number of pixels); E - Maximum difference of the current and previous $M_{x,y}(O)$ (in number of pixels); F - Margin for overlapping with other objects' bounding boxes (in number of pixels).}\label{Table I}
			\begin{tabular}{l|cccccc} \hline\hline
			{\bf Scenario} & {\bf A} & {\bf B} & {\bf C} & {\bf D} & {\bf E} & {\bf F} 
			\\ 
				\hline
				{\bf Scenario 1} & 50 $\times$ 50 pixels & 25 & 300 & 0 & 17 & 7\\
				{\bf Scenario 2} & 50 $\times$ 50 pixels & 22 & 450 & 6 & 30 & 20\\
				{\bf Scenario 3} & 40 $\times$ 40 pixels & 30 & 600 & 0 & 30 & 20\\
				{\bf Scenario 4} & 40 $\times$ 40 pixels & 30 & 450 & 0 & 30 & 20\\
			\hline\hline 
			\end{tabular}
		
	\end{table}

To track multiple objects, we colored the outline of~$b$ as white once~$b$ detects an object. This way, the system will not keep detecting and tracking the same object within the same~$f$.

\subsection{Data Computation}
\subsubsection{Tracing Trajectories and Computing Distances}
We traced the trajectories or the path traveled of each pedestrian in the image using each object's $M_{x,y}(O)$ for every~$f$. Given the coordinates of the $(x,y)$ in each~$f$, we used the cononical two-dimensional distance formula, the $L_2$ metric, to compute for the distance~$d_p(O)$ traveled by each pedestrian per~$f$. 
\subsubsection{Computing Velocities and Accelerations}
	Given the change in the position ($\delta d_p(O)$) over the change in time ($\delta t$) of each tracked head, we obtained the instantaneous velocity $v_p(O)$ at each~$f$ of each tracked pedestrian by a simple ratio $\delta d_p(O) / \delta t_p$. We computed the  acceleration~$a_p(O)$ of each tracked pedestrian by the change in velocity over time ($\delta v_p(O)/\delta t$).
	
\subsection{Presentation of Measured Values}
\begin{figure*}[htb]
		\centering
			\epsfig{file=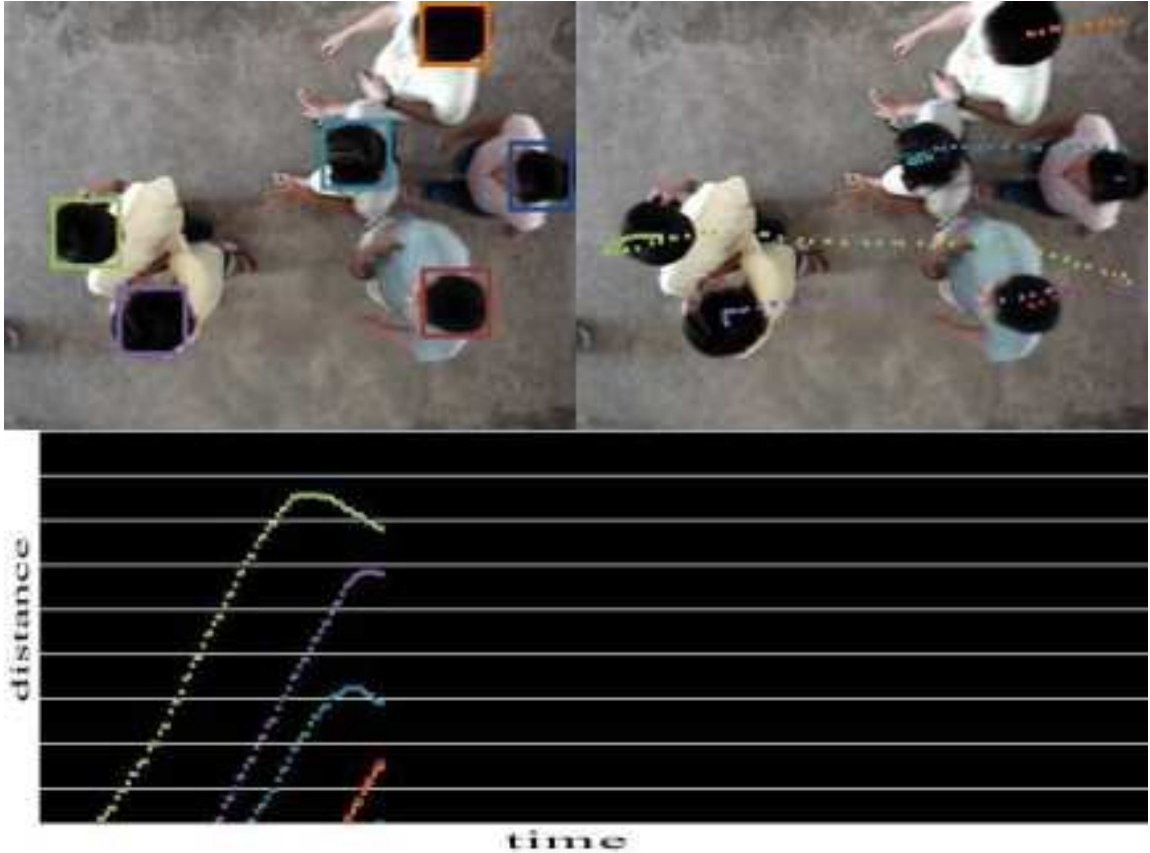,width=6in}
		\caption{An example output frame of the system. The upper Left part contains the input image with trackers (bounding boxes). The upper Right part contains the trajectory of the pedestrians. The lower part contains the distance vs. time graph of the trajectory. This figure is in color in the electronic copy of this paper.}
		\label{Fig3}
	\end{figure*}

We presented the trajectories of the pedestrians by superimposing colored lines following the center of mass $M_{x,y}$ on the video scene (Figure~\ref{Fig3}). We displayed the other motion characteristics of pedestrians as line graphs over time: distance vs. time, velocity vs. time, and acceleration vs. time.
	
\section{Results and Discussion}
In comparing the motion characteristics of the pedestrians at different scenarios, we used the trajectories (superimposed on video sequences) and the graphs generated by the system. 
	
\subsection{Scenario 1}
This scenario shows a freely flowing pedestrians. As captured in the output graphs (Figure~\ref{Figure 5}), uniform speed is shown for all the pedestrians. The small variation shown in the graph can be attributed to the integer arithmetic when dealing with pixel positions and not with the small motion by the pedestrians. However, even if such motion was created by the pedestrians, such variations are small enough to create traffic jams.
	
\subsection{Scenario 2}
In this scenario, the traffic flow of simultaneous crossing of pedestrians from opposite directions is smooth, as captured by the velocity vs. time graph in  Figure~\ref{Figure 6}. Although there were two directions of pedestrian flow, the individuals were moving in uniform lanes. We observed in their respective trajectories (Figure~\ref{Figure 6}~(b)) that there were two dominant divisions of walking lanes. This observation surely verifies the claim of Helbing~\cite{ref13} that the pedestrians self-organize. Figure~\ref{Figure 6}~(c) shows the separate groups of pedestrians. The line graphs starting in distance zero correspond to the pedestrians coming from the left part of the scenario while the other line graphs correspond to the ones coming from the right. Even if the graphs crossed each other, there were no major changes in the magnitude of the pedestrians' motion characteristics. The pedestrians still created uniform lanes of walking direction to avoid moving delays.
	
	\subsection{Scenario 3}
In this scenario, a pedestrian~$P$ stopped in an area in the scene and went back to the direction he came from (i.e., made a U-turn). It was observed in the graphs (Figure~\ref{Figure 7}) that some pedestrians slowed down when they where intersected by~$P$. Others stopped for a certain range of time. As~$P$  moves away from the other pedestrians, the latter slowly recovered and increased their velocity. It shows here that a pedestrian itself can cause delays in pedestrian traffic.
	
	\begin{figure}[htb]
		\centering
			\epsfig{file=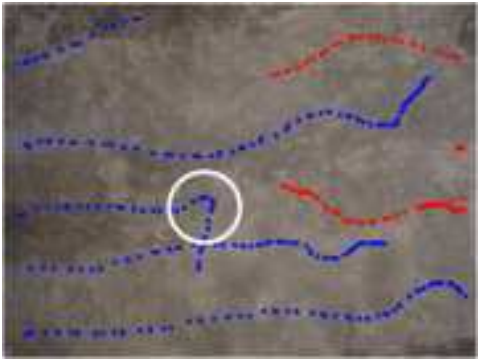,width=2.5in}
		\caption{The encircled path shows the trajectory of a right turn made by a pedestrian to avoid colliding with the incoming traffic. This figure is in color in the electronic copy of this paper.}
		\label{Figure 4}
	\end{figure}	
	
	\subsection{Scenario 4}
In this scenario, a counter flow exhibited slow down of motion of the pedestrians before and during the intersection. We observed no self-organization among pedestrians here. Instead, the individuals moved to different directions just to reach their destination and to avoid colliding with other pedestrians. We observed from some of the pedestrians' trajectories that they make turning motion to avoid the obstacles along their way (see Figure~\ref{Figure 4}). Even with this obstacle avoidance behavior, because everybody is avoiding the counter flow, a short-term traffic jam that lasted for a short time was observed. When a pedestrian moved out of the intersection, the pedestrian was able to get back to a uniform speed, as evident in Figure~\ref{Figure 8}.

\section{Conclusion}

In this study, we created a machine vision system that automatically captures the dynamics of pedestrian traffic. The system was able to detect and track multiple pedestrians and effectively display some of their motion characteristics. We used the system to analyze and quantify the behavior of the pedestrian in four different traffic scenarios, namely:
\begin{enumerate}
  \item {\bf Scenario 1}: Pedestrians were flowing smoothly in one direction (left-to-right, LR).
	\item {\bf Scenario 2}: Pedestrians were flowing in two opposite directions simultaneously (LR and right-to-left, RL).
	\item {\bf Scenario 3}: Pedestrians were initially flowing from a single direction (LR) and then suddenly, a pedestrian stops in the middle and makes a U-turn.
	\item {\bf Scenario 4}: Pedestrians were moving against counter flows: They were passing by each other in several and inconsistent pathways or lanes.
\end{enumerate}
		
Using the measured metrics by the system, the results of our analysis suggest that pedestrians themselves alter the traffic condition. The motion characteristics of a pedestrian depend on its desired magnitude and on the delays in its environment. In counter flows (as in Scenario 4), if pedestrians self-organize, the traffic flow can be smooth, otherwise a traffic jam may build up.
	
By learning the behavior of pedestrians in different traffic scenarios, we could efficiently devise safety and security measures in crowded areas, avoid accident, injuries, and even loss of lives. This system will be helpful in the field of architecture for the structural designs of pedestrian hallways and pathways in buildings. We recommend this system for use in the study of pedestrian evacuation during fire escaping and for avoidance accidents such as stampedes.

\section{Acknowledgments}
We value the unselfish technical inputs of Dr. Vladimir Y. Mariano, Associate Professor of the Institute of Computer Science, UPLB, in the area of image processing and video analysis. We also thank the Institute of Computer Science, UPLB, for the use of its computing equipments to create, and test the system. Lastly, we thank the unnamed pedestrians that we captured in our videos. Unknown to them, they had been willing experimental subjects in our study. 

\bibliographystyle{plain}
\bibliography{pedestrian}

\begin{thebibliography}{10}

\bibitem{ref13}
D.~Helbing (2).
\newblock Safety management at large events: The problem of crowd panic, 2006.

\bibitem{ref12}
T.~Arns.
\newblock Video films of pedestrian crowds, 1993.

\bibitem{ref3}
V.~J. Blue.
\newblock Cellular automata microsimulation for modeling bi-directional
  pedestrian walkways.
\newblock {\em Transportation Research Part B: Methodological}, 35(3):293--312,
  2001.

\bibitem{ref9}
J.~Ganem.
\newblock A behavioral demonstration of {F}ermat's principle.
\newblock {\em The Physics Teacher}, 36, 1998.

\bibitem{ref17}
I.~Griffiths.
\newblock {\em Principles of Biomechanics and Motion Analysis}.
\newblock Lippincott Williams and Wilkins, USA, 2006.

\bibitem{ref5}
D.~Helbing.
\newblock A fluid dynamic model for the movement of pedestrians.
\newblock {\em Complex Systems}, 6:391, 1992.

\bibitem{ref11}
D.~Helbing.
\newblock Traffic and related self-driven many-particle systems.
\newblock {\em Reviews of Modern Physics}, 73:1067, 2001.

\bibitem{ref2}
D.~Helbing, I.~Farkas, and T.~Vicsek.
\newblock Simulating dynamical features of escape panic.
\newblock {\em Nature}, 407(6803):487--490, 2000.

\bibitem{ref6}
D.~Helbing and P.~Molnar.
\newblock Social force model for pedestrian dynamics.
\newblock {\em Physical Review E}, 51(5):4282+, 1995.

\bibitem{ref8}
D.~Helbing, P.~Molnar, I.~Farkas, and K.~Bolay.
\newblock Self-organizing pedestrian movement.
\newblock {\em Environment and Planning B: Planning and Design}, 28:361--383,
  2001.

\bibitem{ref1}
J.~A. Kirkland and A.~A. Maciejewski.
\newblock A simulation attempts to influence crowd dynamics.
\newblock In {\em SMC '03 Conference proceedings : 2003 IEEE International
  Conference on Systems Man and Cybernetics}, pages 4328--4333, Washington,
  D.C., 2003.

\bibitem{ref4}
C.~W. Reynolds.
\newblock Flocks, herds, and schools: A distributed behavioral model.
\newblock {\em Computer Graphics}, 21(4):25--34, 1987.

\bibitem{ref15}
M.~Rowan and F.~Maire.
\newblock An efficient multiple object vision tracking system using bipartite
  graph matching.
\newblock Technical report, School of Software Engineering and Data
  Communication, Queensland University of Technology, Australia, 2006.

\bibitem{ref16}
M.~Sonka, V.~Hlavac, and R.~Boyle.
\newblock Image processing, analysis, and machine vision third edition, 2008.

\bibitem{virtualdub}
{virtualdub.org}.
\newblock {VirtualDub} version 1.6.16, 2009.
\newblock http://www.virtualdub.org.

\bibitem{ref10}
U.~Weidmann.
\newblock Transportation technique for pedestrians, 1993.

\end{thebibliography}

\section*{About the Authors}
Louie Vincent A. Ngoho is an undergraduate student at UPLB. He is a member of the Red Cross Youth of UPLB and of the UPLB DOST Scholars' Society.

Jaderick P. Pabico is an Associate Professor of the Institute of Computer Science, UPLB. He is one the {\bf 2008 Ten Outstanding Young Scientists of the Philippines}. His research interests are in the areas of high-performance computing, intelligent computing, complex systems, computer security and forensics, simulation and modeling, and game programming.

\begin{figure*}
\centering
			\epsfig{file=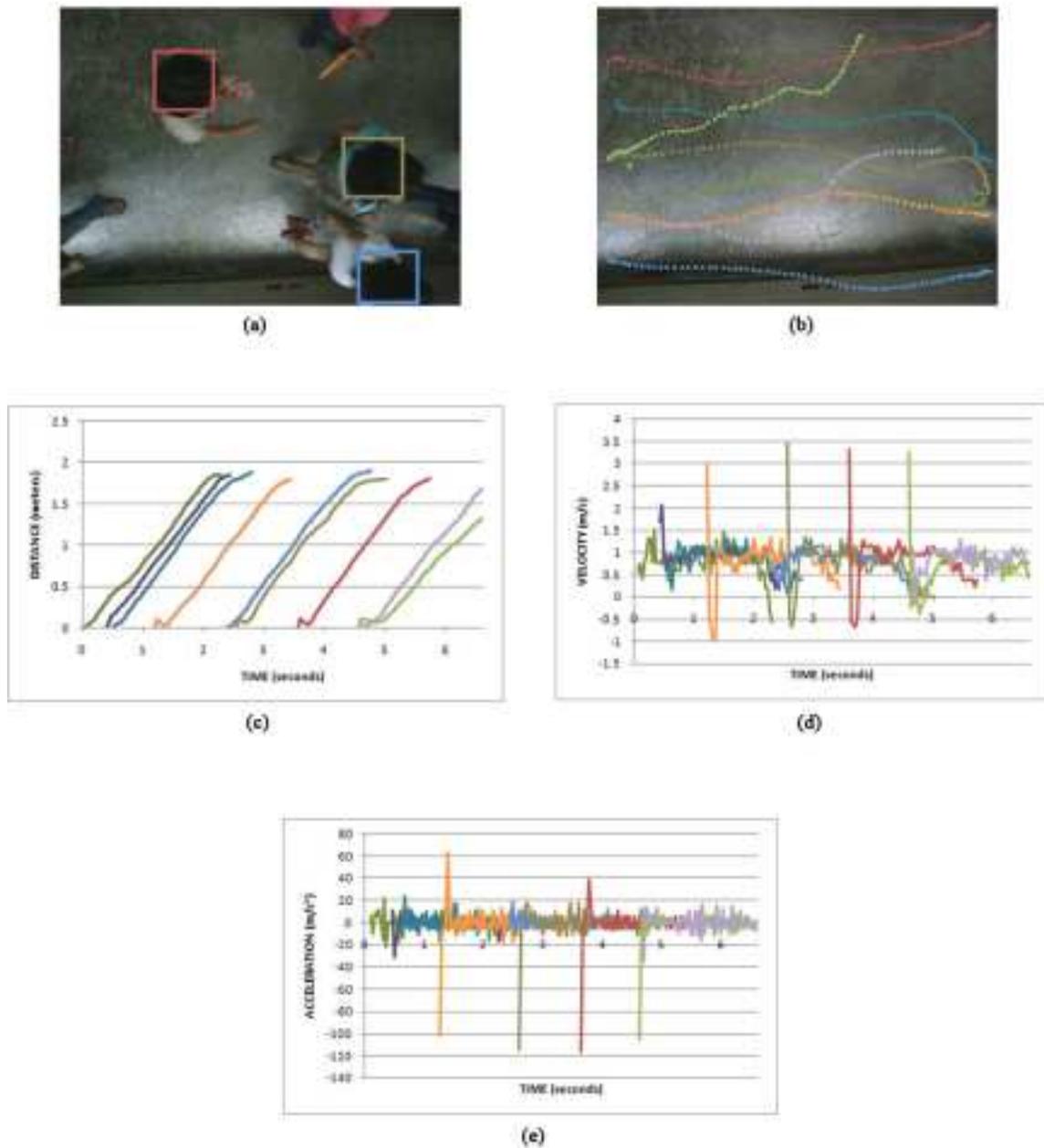,width=6in}
		\caption{(a) An output image, (b) the corresponding trajectories of the pedestrians, (c) the distance vs time graph, (d) the velocity vs time graph, and (e) the acceleration vs time graph for the pedestrians in {\bf Scenario 1}. Each colored set of points in a graph corresponds to the motion characteristic of a single pedestrian. This figure is in color in the electronic copy of this paper.}
		\label{Figure 5}
\end{figure*}

\begin{figure*}
\centering
			\epsfig{file=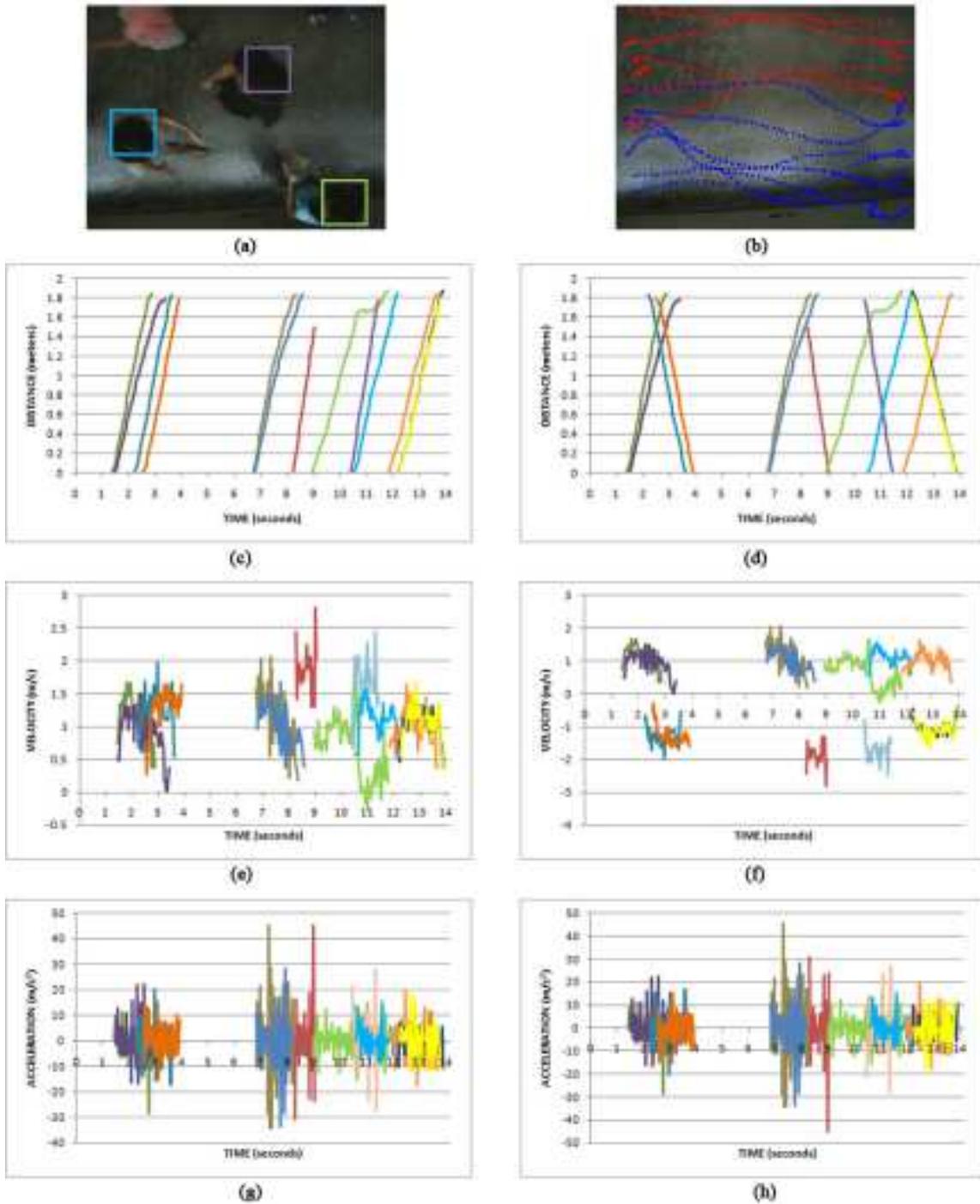,width=6in}
		\caption{(a) An output image, (b) the corresponding trajectories of the pedestrians, (c,d) the distance vs time graphs, (e,f) the velocity vs time graphs, and (g,h) the acceleration vs time graph of the scenario with simultaneous flow of pedestrians in both directions. The trajectory (b) shows separate groups. The red paths represent that of the pedestrians from the right and the blue ones represent pedestrians from the left. The graphs on the right (d, f, h) emphasizes the intersection of motion characteristics of two groups of pedestrians. Graphs of pedestrians from the opposite direction do not start in distance zero to identify separate groups in the graph. This figure is in color in the electronic copy of this paper.}
		\label{Figure 6}
\end{figure*}

\begin{figure*}
\centering
			\epsfig{file=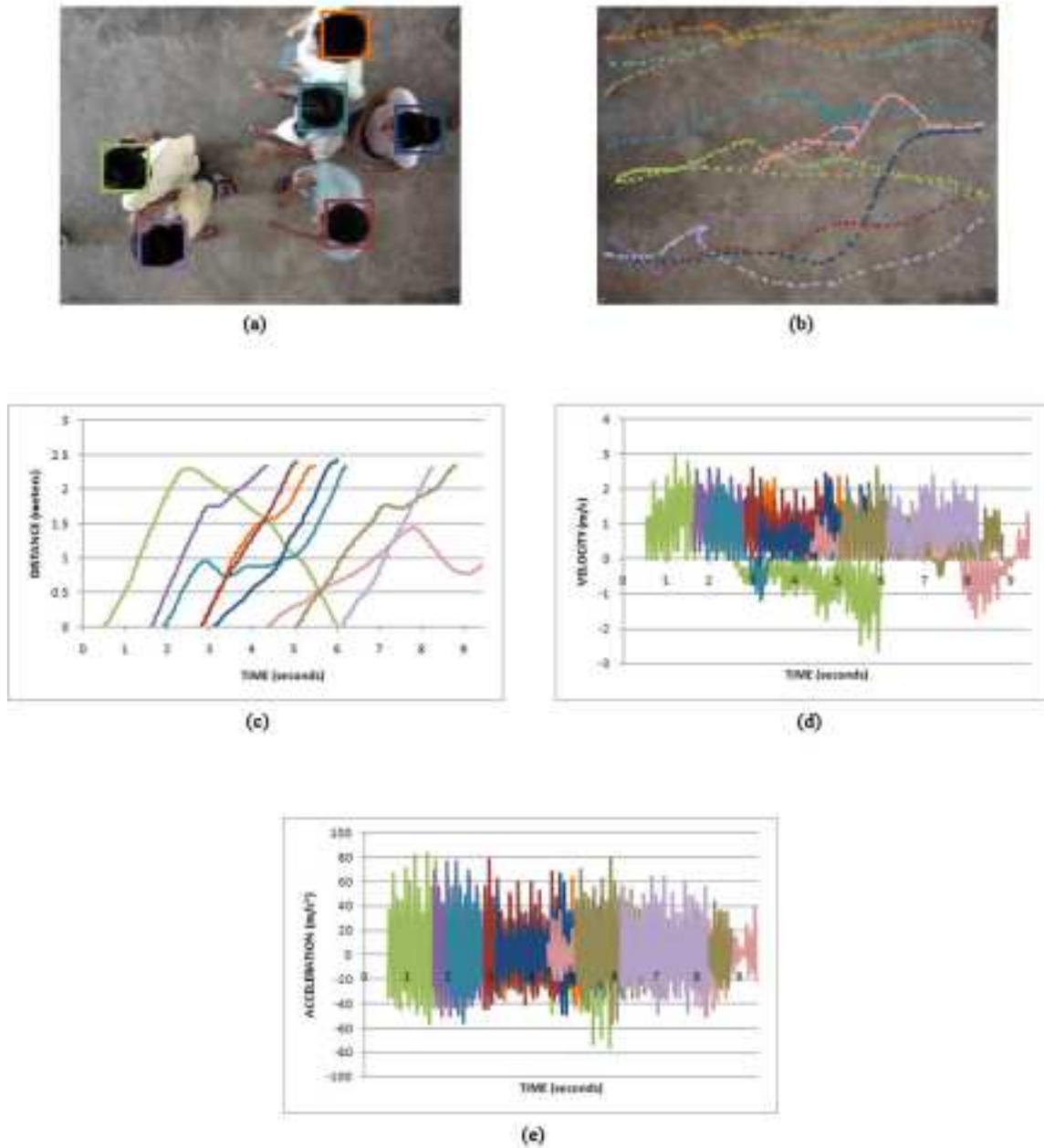,width=6in}
		\caption{(a) An output image, (b) the trajectories of the pedestrians, (c) the distance vs time graph, (d) the velocity vs time graph, and (e) the acceleration vs time graph of the scenario wherein an individual suddenly stops and does a U-turn. This figure is in color in the electronic copy of this paper.}
		\label{Figure 7}
\end{figure*}

\begin{figure*}
\centering
			\epsfig{file=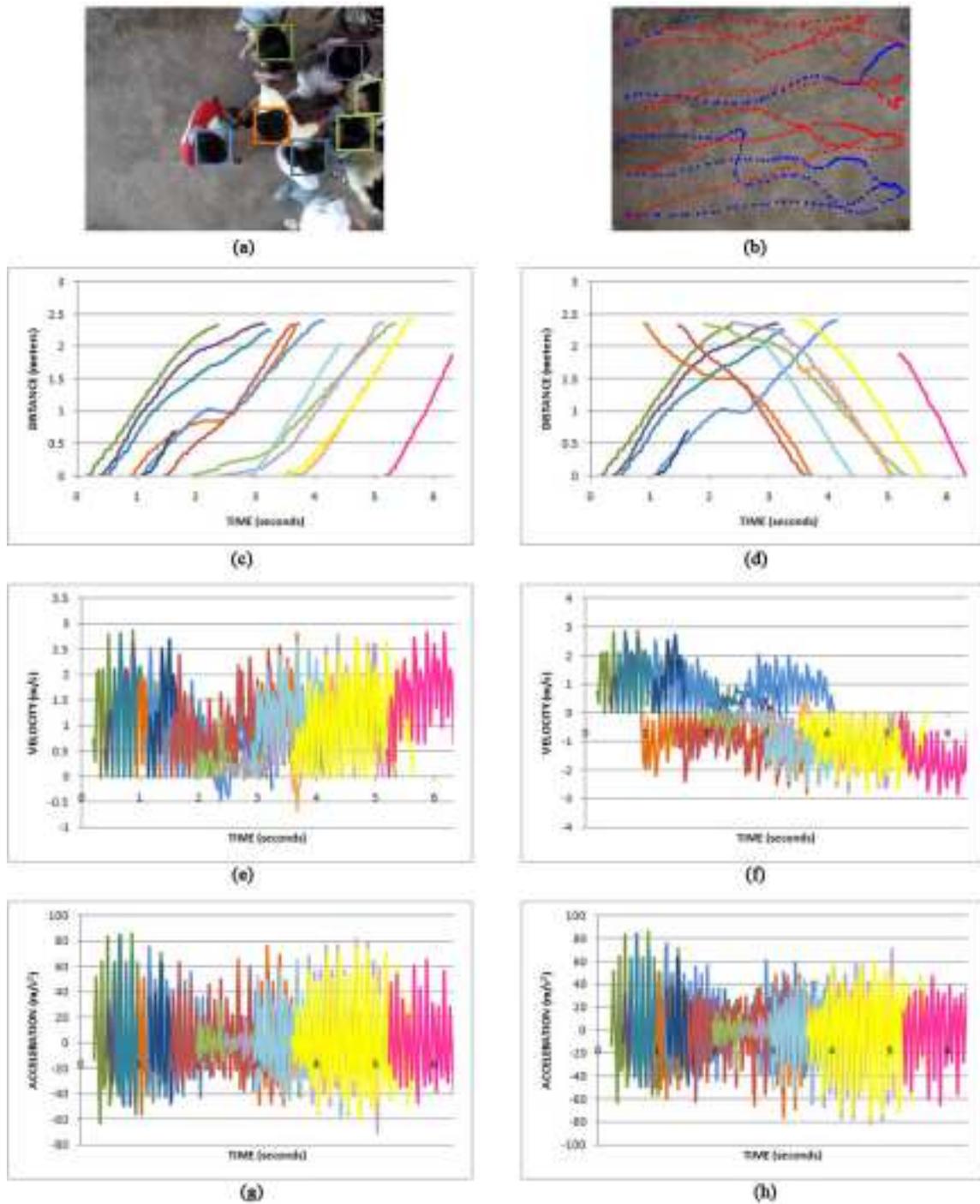,width=6in}
		\caption{(a) An output image, (b) the trajectories of the pedestrians, (c,d) the distance vs time graphs, (e,f) the velocity vs time graphs, and (g,h) the acceleration vs time graphs of the counter flow scenario. The trajectory (b) shows separate groups. The red paths represent that of the pedestrians from the right and the blue ones represent pedestrians from the left. The graphs on the right (d, f, h) emphasizes the intersection of motion characteristics of two groups of pedestrians. Graphs of pedestrians from the opposite direction do not start in distance zero to identify separate groups in the graph. This figure is in color in the electronic copy of this paper.}
		\label{Figure 8}
\end{figure*}

\begin{figure*}
\centering
    \epsfig{file=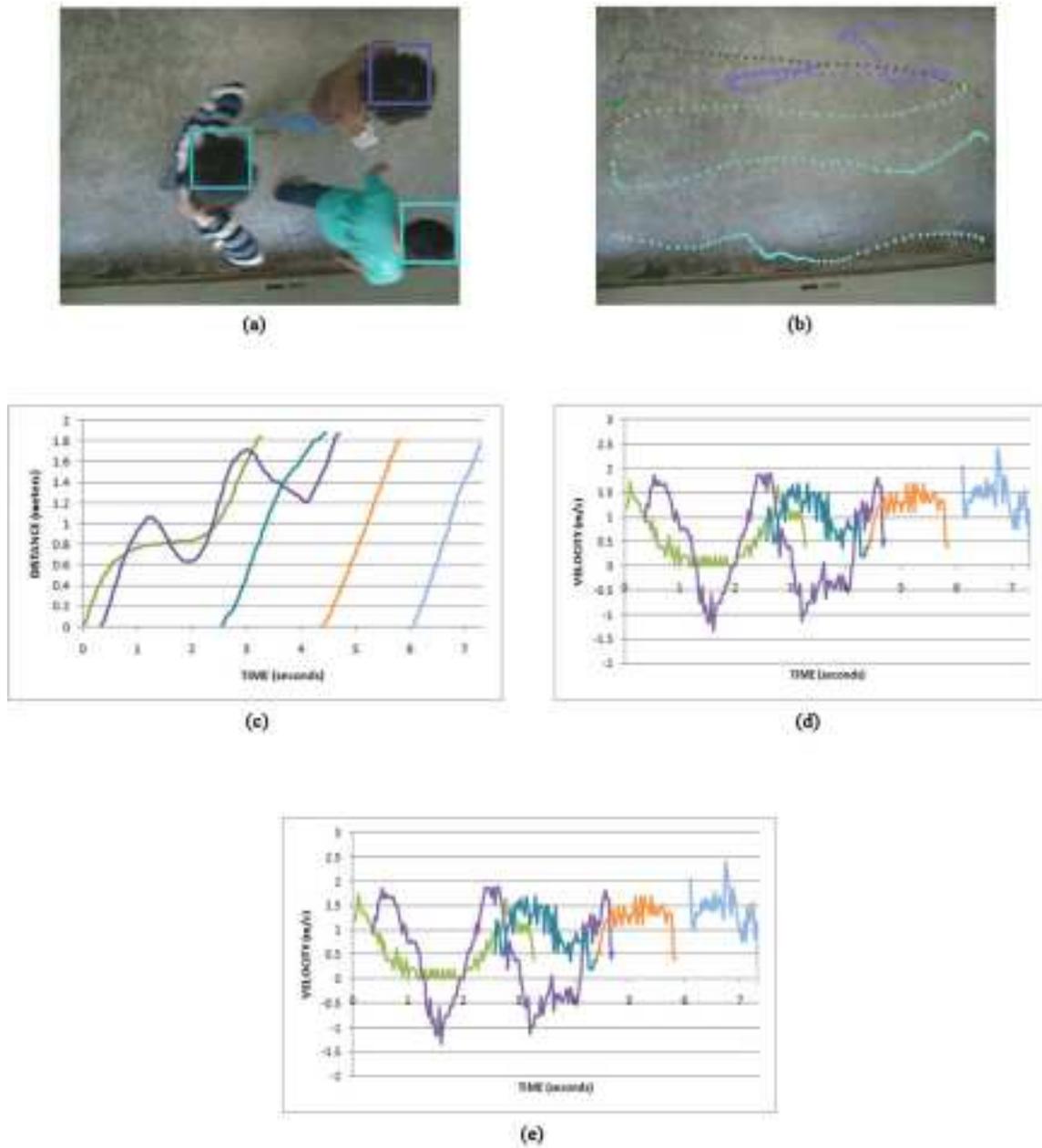,width=6in}
		\caption{(a) An output image, (b) the trajectory of the pedestrians, (c) the distance vs time graph, (d) the velocity vs time graph, and (e) the acceleration vs time graph of the scenario wherein a pedestrian moves back and forth, showing variations in his motion characteristics. This figure is in color in the electronic copy of this paper.}
		\label{Figure 9}
\end{figure*}

\end{document}